\pdfoutput=1
\documentclass[10pt,twocolumn,letterpaper]{article}

\usepackage[pagenumbers]{cvpr} %

\usepackage{cuted} %

\usepackage{currfile} %

\usepackage{caption} %

\usepackage{lipsum}

\usepackage{multirow}
\newcommand{\spd}[1]{#1$\times$}   %

\usepackage{fontawesome5} %
\usepackage{twemojis}     %
\newcommand{\hficon}{\raisebox{-.2ex}{\twemoji[height=1.25\fontcharht\font`X]{1f917}}} %

\definecolor{cvprblue}{rgb}{0.21,0.49,0.74}
\usepackage[pagebackref,breaklinks,colorlinks,allcolors=cvprblue]{hyperref}

\def\confName{CVPR}
\def\confYear{2027}

\title{MMFlash: Faster Speculative Decoding for Vision-Language Models}

\author{
Sicheng Feng\quad
Zigeng Chen\quad
Song Wang\quad
Gongfan Fang\quad
Xinchao Wang\thanks{Corresponding author.}\\
National University of Singapore\\
{\tt\small fengsicheng@u.nus.edu, xinchao@nus.edu.sg}\\[2mm]
{\normalsize \faGithub~Code: \href{https://github.com/fscdc/MMFlash}{MMFlash}\qquad
\hficon~Models: \href{https://huggingface.co/collections/FSCCS/mmflash}{MMFlash}}
}

\begin{document}
\maketitle
\makeatletter\global\@topnum\z@\makeatother
\begin{figure}[t]
\centering
\includegraphics[width=\linewidth]{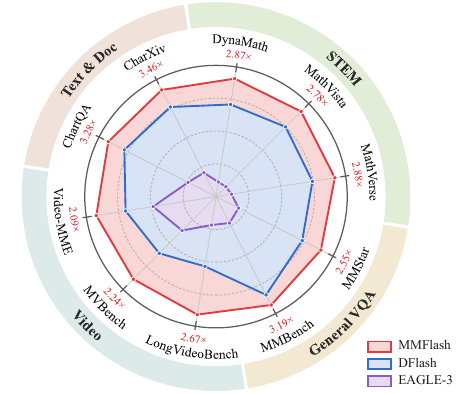}  \\
\vspace{-2mm}
\caption{Decoding speedup comparison with SOTA speculative decoding methods (EAGLE-3 \& DFlash) on SGLang.}
\label{fig:teaser}
\vspace{-5mm}
\end{figure}

\begin{figure*}[t]
\centering
\includegraphics[width=\linewidth]{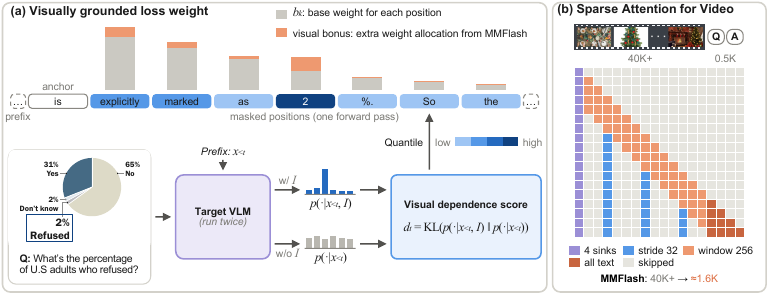}  \\
\vspace{-2mm}
\caption{Overview of MMFlash. (a)~The target VLM is run twice on the same prefix $x_{<t}$, with and without the image $I$, and the visual dependence score $d_t$ is the KL divergence between the two next-token distributions. Each masked position of a block keeps its base weight $b_k$ and gains a visual bonus that depends on $d_t$. (b)~The MMFlash drafter adopts a simple sparse attention pattern for video inputs, shrinking a context of more than 40K tokens to about 1.6K.} %
\label{fig:overview}
\vspace{-4mm}
\end{figure*}

\begin{abstract}
Speculative decoding accelerates generation by letting a lightweight drafter propose tokens for parallel verification by a target model. 
Diffusion drafters such as DFlash propose an entire token block in a single forward pass, but mainly focus on text-only tasks, leaving multimodal drafting underexplored. 
To study this setting, we first introduce a metric of token-level visual dependence based on the KL divergence between next-token distributions with and without visual input. 
We find that acceptance rates decrease from 46\% in the lowest visual-dependence quartile to 22\% in the highest, with this disparity persisting after multimodal training.
Motivated by this observation, we propose MMFlash, a simple and effective framework that adaptively upweights visually dependent tokens that the drafter predicts poorly.
For long video inputs, MMFlash further employs sparse attention to reduce redundant attention computation and improve drafting quality.
Across various multimodal benchmarks, MMFlash achieves lossless decoding speedups of up to $3.58\times$, outperforming EAGLE-3 and DFlash by up to 195.4\% and 17.1\%, respectively.
\end{abstract}

\section{Introduction}
\label{sec:intro}

Vision-language models (VLMs)~\citep{achiam2023gpt,bai2025qwen3,team2026gemma,wang2025internvl3} have demonstrated remarkable capabilities across multimodal tasks such as visual question answering (VQA)~\citep{antol2015vqa,wang2024measuring,yue2024mmmu}, video understanding~\citep{wu2024longvideobench,li2024mvbench,fu2026labinstruct}, and visual reasoning~\citep{liu2025visual,feng2026reasonmap,xu2026visulogic,feng2026rewardmap}. 
However, their inference efficiency remains constrained by autoregressive generation, which produces tokens sequentially. This bottleneck is further exacerbated by the growing adoption of reasoning models that generate longer responses~\citep{jaech2024openai,guo2025deepseek,feng2025efficient}. 
Speculative decoding~\citep{leviathan2023fast,chen2023accelerating,xia2024unlocking} has emerged as a prominent solution: a small model drafts multiple tokens, which a larger target model then verifies in parallel, substantially accelerating decoding.

The performance of speculative decoding largely depends on draft quality and drafting cost. Early approaches include Medusa~\citep{cai2024medusa}, which augments the target model with multiple decoding heads, and the EAGLE series~\citep{li2024eagle1,li2024eagle2,li2026eagle,lieagle4}, which drafts from target model features and uses tree-based verification to increase acceptance lengths. For autoregressive drafters, however, each drafting step requires a forward pass, making drafting cost grow linearly with draft length. This favors lightweight architectures, often at the expense of draft quality. 
Inspired by diffusion language models (DLMs)~\citep{nie2025large,ye2025dream,chen2026dmax}, recent methods~\citep{liu2026dart,huang2026domino,wu2026d,zhang2026dspine} such as DFlash~\citep{chen2026dflash} use block diffusion~\citep{arriola2025block} to generate an entire draft block in a single forward pass, decoupling the number of sequential drafting steps from the number of proposed tokens. The computational budget previously spent on repeated sequential passes can thus support deeper, higher-capacity drafters, improving draft quality. 
Building on these advances, we extend diffusion drafting to multimodal models, a direction that remains largely unexplored.

To better extend to the vision field, we first introduce a token-level measure of visual dependence: the KL divergence between the target model’s distributions with and without visual input, conditioned on the same textual prefix. 
Using this metric, we analyze two DFlash drafters for Qwen3.5-4B (Figure~\ref{fig:motivation}): the official text-trained drafter and a counterpart trained on multimodal data following the official recipe. Acceptance rates decrease monotonically with visual dependence, from 46\% in the lowest quartile to 22\% in the highest. Moreover, the highest quartile accounts for 44\% of first rejections within draft blocks, compared with only 7\% for the lowest. 
This disparity persists after training with multimodal data.

To address this challenge, we propose MMFlash (Figure~\ref{fig:overview}), a simple and effective framework for VLMs. Its core is a visually aware loss reweighting scheme that uses the above visual dependence scores to emphasize visually dependent tokens in the DFlash cross-entropy objective. 
The additional weight scales with the drafter’s prediction error probability, emphasizing tokens it has yet to learn and gradually returning to the baseline weight as predictions improve. 
For long videos, drafter attention concentrates on nearby positions, with substantial redundancy among visual tokens. We thus equip the MMFlash drafter with a simple sparse attention pattern that retains a few attention sinks, strided visual tokens, all text tokens, and a local window, consistently across training and inference.

We evaluate MMFlash primarily on Qwen3.5 series~\citep{qwen3.5} across seven image benchmarks covering text and document understanding, general VQA, and STEM reasoning, as well as three long video benchmarks (Figure~\ref{fig:teaser}). Baselines include native MTP, EAGLE-3, and DFlash. 
On image benchmarks, MMFlash achieves the longest average acceptance lengths and highest decoding speedups across all settings. Under greedy decoding, average speedups reach 3.00-3.05$\times$, exceeding DFlash by 10.3-10.9\%. On long video benchmarks, sparse attention combined with minor video fine-tuning yields average speedups of 2.33-2.71$\times$, outperforming DFlash by 11.2-23.2\%. 
MMFlash remains effective under high concurrency and generalizes to other model families and larger scales.

Overall, we introduce MMFlash, a novel diffusion drafting framework for speculative decoding in VLMs. By prioritizing poorly predicted, visually dependent tokens during training, MMFlash addresses a key bottleneck in draft acceptance. Comprehensive experiments demonstrate its effectiveness and generalizability, establishing MMFlash as a strong baseline for speculative decoding in VLMs.

\section{Related Work}
\label{sec:related}

We briefly review the most relevant work here and provide a more detailed discussion in Appendix~\ref{supp:related}.

\noindent \textbf{Speculative Decoding for VLMs.}
Speculative decoding~\citep{leviathan2023fast,chen2023accelerating} lets a lightweight drafter propose tokens that the target model verifies in parallel, leaving its output distribution unchanged and thus achieving lossless acceleration. Drafters range from decoding heads~\citep{cai2024medusa} to feature-level drafters with tree-structured verification~\citep{li2024eagle1,li2024eagle2,li2026eagle,lieagle4}. 
For VLMs, existing methods differ mainly in how the drafter handles visual input: it receives the full visual input~\citep{lin2025speculative,huo2025spec,ganesan2025massv,tong2026sage}, only distilled visual information~\citep{hu2026dream,kang2026vispec,zhang2025sparse,ji2025specvlm,lv2026hippo,kong2026parallelvlm}, or no visual tokens at all~\citep{gagrani2024speculative,xie2026hivis,zhang2026sparrow}, and some methods switch among these strategies adaptively~\citep{huang2025specvlm,lee2026tabed,lee2024batch}. MMFlash follows the third and most efficient paradigm, since the target has already fused the visual content into the hidden states that the drafter reads. Orthogonal efforts relax verification~\citep{ji2026see}, speculate at the agent level~\citep{huang2026speceyes}, or specialize in document OCR~\citep{liao2026hsd}.

\noindent \textbf{Speculative Decoding with Diffusion Drafters.}
Diffusion language models (DLMs)~\citep{sahoo2024simple,zheng2024masked,nie2026large,ye2025dream,khanna2025mercury,song2025seed} generate text by denoising many positions in parallel, and block diffusion~\citep{arriola2025block} applies this denoising within blocks that are generated autoregressively. Since a DLM can propose a whole block in one forward pass, early work explores diffusion-style drafting by mixing diffusion and autoregressive objectives in one model~\citep{liu2511tidar}, adapting autoregressive large language models (LLMs) for parallel drafting~\citep{samragh2025your}, or drafting with large pre-trained DLMs~\citep{li2026diffuspec,sandler2026specdiff}. DFlash~\citep{chen2026dflash} makes diffusion drafting simple and effective with a lightweight drafter that emits an entire block in one forward pass, and follow-up work explores related designs~\citep{liu2026dart,huang2026domino,fu2026nemotron} and dynamic position-wise loss weights~\citep{wu2026d,gu2026verification,yang2026spec}.

\section{Preliminaries}
\label{sec:prelim}

\begin{figure*}[t]
\centering
\includegraphics[width=\linewidth]{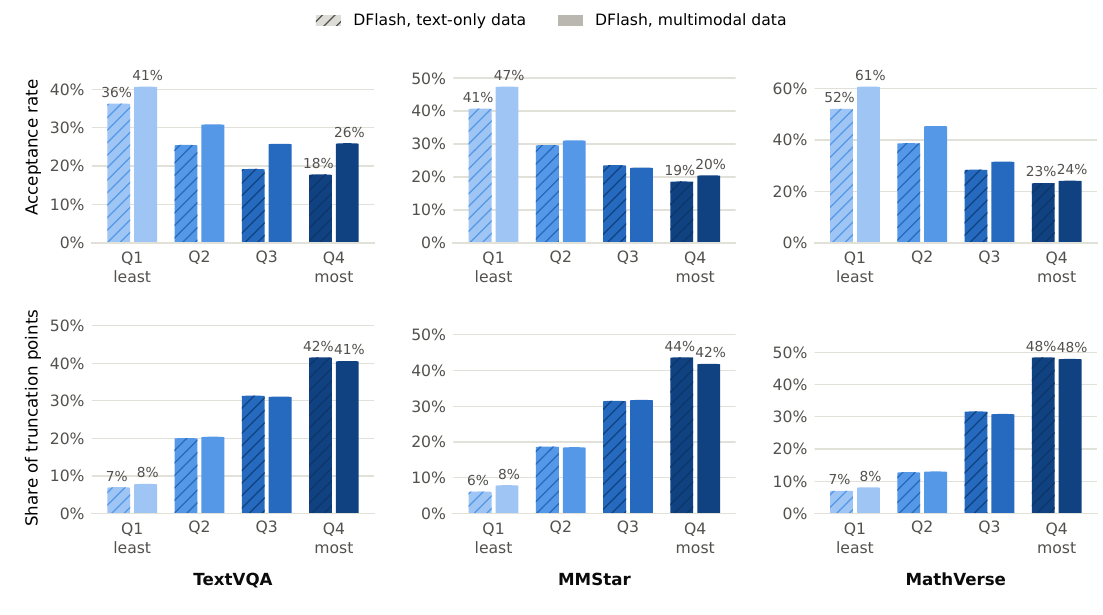}  \\
\vspace{-6mm}
\caption{Empirical studies on Qwen3.5-4B with three benchmarks. For each benchmark, we randomly select 200 samples and sample up to 128 anchor positions per sample. \textbf{Top}: acceptance rate of drafted tokens. \textbf{Bottom}: share of truncation points falling in each quartile. Visual dependence increases from Q1 to Q4.} %
\label{fig:motivation}
\vspace{-4mm}
\end{figure*}

As a baseline for our method, we briefly introduce DFlash~\citep{chen2026dflash}, highlighting its inherent advantages for speculative decoding. Unlike autoregressive drafters, which require multiple sequential forward passes to propose draft tokens, DFlash generates an entire draft block in a single forward pass. This parallelism decouples the number of sequential drafting passes from the number of draft tokens, allowing the computational budget otherwise spent on the drafting passes to support a deeper, higher-capacity drafter.

Concretely, at each drafting step, DFlash conditions on context features extracted from the target model and processes a block of $B$ positions, one anchor token followed by $B-1$ mask tokens, filling all masks in parallel. During training, anchors are randomly sampled from a response generated by the target model, and the $B-1$ tokens after each anchor are masked for reconstruction. This mirrors inference, where each block starts with the bonus token from the preceding verification step. Let $z^\star_k$ denote the token that the target decoding policy selects at the $k$-th masked position, and let $q_k$ be the probability that the drafter assigns to it given the prefix. Training reduces to a cross-entropy fit to the target's own choices, except that the $B-1$ positions do not contribute equally to the loss:
\begin{equation*}
  \mathcal{L}_{\mathrm{DFlash}} = \sum_{k=1}^{B-1} w_k \bigl(-\log q_k\bigr),
  w_k = \exp\!\left(-\frac{k-1}{\gamma}\right),
\end{equation*}
where the hyperparameter $\gamma > 0$ sets how fast the weight decays along the block. This decay mirrors how a block is consumed at inference time. Verification accepts a prefix of the block, so a rejection at position $k$ discards every token after it, and position $k$ lengthens the accepted sequence only when positions $1, \dots, k-1$ have all been accepted.

\section{Method}
\label{sec:method}

In this section, we first identify a key limitation of extending DFlash to multimodal speculative decoding, which motivates us to propose our MMFlash. Then, we introduce our MMFlash in detail.

\subsection{Motivation}
\label{sec:motivation}

DFlash has been thoroughly validated in the text setting. Since our target is VLMs on multimodal data, the natural first question is how DFlash behaves on tokens that depend on the multimodal inputs (\textit{e.g.}, images).
Answering it requires a per-token measure of visual dependence, and we take this measure from the target model rather than from the drafter or from attention maps. For each generated token, we run the target model twice, once with the image and once with the image removed while keeping the text prompt and the generated prefix fixed, and define the token's visual dependence as the KL divergence between the two resulting next-token distributions:
\begin{equation}
  d = \mathrm{KL}\bigl( p(\cdot \mid \text{prefix}, I) \,\|\, p(\cdot \mid \text{prefix}) \bigr),
  \label{eq:vdep}
\end{equation}
where $p$ denotes the next-token distribution of the target model and $I$ the input image.
This choice follows directly from how speculative decoding works. The sole objective of the drafter is to imitate the distribution of the target model, and a draft token is accepted only when it matches the target model's own prediction.
What matters is therefore how much of the prediction that the drafter must reproduce is contributed by the image, which is exactly what the divergence above measures.

Then, we conduct an empirical study on Qwen3.5-4B with two DFlash drafters: (1) the officially released drafter\footnote{\url{https://huggingface.co/z-lab/Qwen3.5-4B-DFlash}}, trained on text-only data; and (2) a drafter we train following the official recipe on roughly 900K multimodal samples drawn from LLaVA-OV-1.5-Instruct-Data\footnote{\url{https://huggingface.co/datasets/mvp-lab/LLaVA-OneVision-1.5-Instruct-Data}}. Every drafted token is scored by the visual dependence defined above and assigned to a quartile from Q1 (least visual-dependent) to Q4 (most). We then report two statistics per quartile: the acceptance rate of drafted tokens; and the share of truncation points that fall in each quartile, where the truncation point of a block is the first token at which the drafter's prediction disagrees with the target's. As shown in Figure~\ref{fig:motivation}, across benchmarks and drafters the acceptance rate falls monotonically with visual dependence (46\% in Q1 vs.\ 22\% in Q4 on average, a 2.1$\times$ gap), while truncation points pile up at the image-dependent end: Q4 holds 44\% of them against 7\% in Q1, and Q3-Q4 together account for 72-80\%. We further report the acceptance rate of drafted blocks grouped by the visual dependence of their first predicted token in Appendix~\ref{supp:empirical_study} (Figure~\ref{fig:motivation_supp}). The block-level view shows the same trend: a block that opens on an image-dependent token retains far less of its draft, with acceptance falling from 33\% in Q1 to 14\% in Q4 on average (a 2.4$\times$ gap).
Overall, these results indicate that the acceptance loss of DFlash on vision-language targets concentrates on visually dependent tokens and that training on multimodal data alone does not close this gap. This observation motivates MMFlash, which explicitly directs training toward such tokens rather than treating all positions alike.

\subsection{MMFlash}

The study above shows where a DFlash drafter loses its acceptance under the multimodal setting: on the tokens whose prediction the image decides. DFlash's objective is blind to this. Its weight $w_k$ depends only on the position inside a block, so a token that merely continues a phrase and a token that reads a number off a chart receive the same gradient. Yet the quantity that exposes the gap, the visual dependence $d$, is computed from the target alone, needs no drafter, and can be scored once offline for the training data. MMFlash therefore turns the diagnostic of Section~\ref{sec:motivation} into a training signal: it reweights the block loss so that the drafter spends its capacity where the image matters, and only while it has not yet learned those tokens.

For every token $t$ of a training response, we compute its visual dependence $d_t$ with Equation~\ref{eq:vdep}, \ie, with the target model, the image $I$, and the prefix $x_{<t}$ fixed. The score $d_t$ is precomputed once offline and stored with the training data, incurring no additional training cost. Following the analysis in Section~\ref{sec:motivation}, we likewise bucket the tokens in the training set $\mathcal{D}$ by this score, and denote by $\tilde g_t\in[0,1]$ the quantile of $d_t$ among all tokens in $\mathcal{D}$ (see Appendix~\ref{supp:quantile} for details). Visual dependence tells us where the image matters; it does not tell us whether the target has settled on a prediction there. The drafter is trained to reproduce the target's choice, and extra effort pays off only where there is a definite choice to reproduce: on a token whose with-image distribution is still flat, the target picks among near-tied candidates, and that pick is not predictable from its features no matter how the token is weighted. We therefore scale the score by the target's confidence with the image, measured by the entropy $H_t$ of its with-image distribution $p(\cdot \mid x_{<t}, I)$:
\begin{equation}
  g_t = \tilde g_t \cdot \exp(-H_t).
  \label{eq:gate}
\end{equation}
Hence $g_t\in[0,1]$ is highest where the image both changes the prediction and settles it, which is exactly where additional effort can turn into acceptance.

\begin{figure*}[t]
\centering
\includegraphics[width=\linewidth]{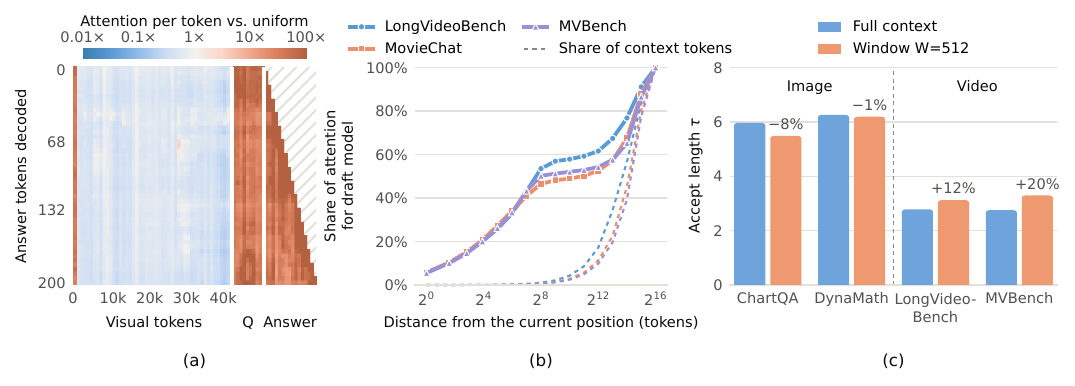}  \\
\vspace{-5mm}
\caption{Empirical studies on Qwen3.5-4B across video benchmarks (60 random samples per benchmark). (a)~Per-token attention of the MMFlash drafter on one LongVideoBench question, relative to
a uniform spread over the visible context; rows are decoding steps. (b)~Cumulative share of draft attention within a given distance of the current position (solid), against the share of context tokens (dashed). (c)~Acceptance length with the full draft context and with a window of the most recent 512 positions.}
\label{fig:sparse_motivation}
\vspace{-4mm}
\end{figure*}

The score says which tokens deserve effort. Two more things decide whether effort at a block position is useful at all: whether verification will ever reach that position, and whether the drafter still has something to learn there. MMFlash therefore keeps the cross-entropy fit of DFlash and replaces its positional weight by a product of three factors. For the $B-1$ masked positions of a block, each position $k$ inherits the static visual grounding $g_k\in[0,1]$ of its target token from Equation~\ref{eq:gate}, which marks whether the token is worth extra effort at all. We combine it with two further factors in the training loss, where $\operatorname{sg}(\cdot)$ denotes the stop-gradient operator:
\begin{equation}
  \begin{aligned}
    w_k &= b_k\,\bigl(1 + \alpha\, g_k\,(1 - \operatorname{sg}(q_k))\bigr), \\
    \mathcal{L}_{\text{MMFlash}} &= \sum_{k=1}^{B-1} w_k\,(-\log q_k),
  \end{aligned}
  \label{eq:mmflash}
\end{equation}
where $b_k$ and $(1-\operatorname{sg}(q_k))$ play complementary roles. $b_k$ is a verification-aware positional weight: since verification accepts only a prefix of the block, the gradient at position $k$ matters only if positions $1,\dots,k$ all survive. We estimate this chain survival from the drafter's own confidence,
\begin{equation}
  b_k = \prod_{i=1}^{k}\bar q_i,
  \qquad
  \bar q_i = (1-\lambda)\,\operatorname{sg}(q_i) + \lambda,
  \label{eq:prefix}
\end{equation}
where the smoothing $\lambda\in[0,1]$ keeps every factor above $\lambda$, so that a single miss weakens the later positions of a block without silencing them. 
Finally, $(1-\operatorname{sg}(q_k))$ adds a \emph{dynamic} component: it is the drafter's current probability of missing the token, so the boost is largest while a visually grounded token is still unlearned and fades to zero as the drafter fits it, shifting the extra effort elsewhere instead of over-fitting visual tokens at the expense of textual ones.

Unless stated otherwise we use $\lambda=0.5$ and $\alpha=1$; further implementation details are given in Appendix~\ref{supp:mmflash_details}.

\begin{table*}[t]
  \centering
  \small
  \setlength{\tabcolsep}{2pt}
  \begin{tabular*}{\linewidth}{@{\extracolsep{\fill}}ll*{16}{c}@{}}
    \toprule
    \multirow{2}{*}{Model} & \multirow{2}{*}{Method}
      & \multicolumn{4}{c}{\textsc{Text \& Doc}} & \multicolumn{4}{c}{\textsc{General VQA}} & \multicolumn{6}{c}{\textsc{STEM}} & \multicolumn{2}{c}{\multirow{2}{*}{\textit{Avg.}}} \\
    \cmidrule(lr){3-6} \cmidrule(lr){7-10} \cmidrule(lr){11-16}
    & & \multicolumn{2}{c}{ChartQA} & \multicolumn{2}{c}{CharXiv}
      & \multicolumn{2}{c}{MMStar} & \multicolumn{2}{c}{MMBench}
      & \multicolumn{2}{c}{DynaMath} & \multicolumn{2}{c}{MathVista} & \multicolumn{2}{c}{MathVerse}
      & \multicolumn{2}{c}{} \\
    \midrule
    \multicolumn{2}{c}{Temperature $= 0$}
      & {\scriptsize $\tau$} & {\scriptsize Speedup} & {\scriptsize $\tau$} & {\scriptsize Speedup} & {\scriptsize $\tau$} & {\scriptsize Speedup}
      & {\scriptsize $\tau$} & {\scriptsize Speedup} & {\scriptsize $\tau$} & {\scriptsize Speedup} & {\scriptsize $\tau$} & {\scriptsize Speedup}
      & {\scriptsize $\tau$} & {\scriptsize Speedup}
      & {\scriptsize $\tau$} & {\scriptsize Speedup} \\
    \midrule
    \multirow{7}{*}{Qwen3.5-4B}
      & MTP~\textcolor{gray}{\scriptsize(3)}  & 3.18 & \spd{1.69} & 3.27 & \spd{1.76} & 2.95 & \spd{1.61} & 2.43 & \spd{1.40} & 3.27 & \spd{1.69} & 3.18 & \spd{1.68} & 3.17 & \spd{1.69} & 3.06 & \spd{1.64} \\
      & MTP~\textcolor{gray}{\scriptsize(7)}  & 4.26 & \spd{1.68} & 4.64 & \spd{1.87} & 3.75 & \spd{1.57} & 2.58 & \spd{1.24} & 4.42 & \spd{1.69} & 4.29 & \spd{1.67} & 4.27 & \spd{1.69} & 4.03 & \spd{1.63} \\
      & MTP~\textcolor{gray}{\scriptsize(15)} & 4.67 & \spd{1.26} & 5.37 & \spd{1.43} & 3.99 & \spd{1.14} & 2.60 & \spd{0.87} & 4.76 & \spd{1.23} & 4.65 & \spd{1.21} & 4.61 & \spd{1.22} & 4.38 & \spd{1.19} \\
      & EAGLE-3~\textcolor{gray}{\scriptsize(16)}  & 4.31 & \spd{1.74} & 4.38 & \spd{1.86} & 3.49 & \spd{1.49} & 2.61 & \spd{1.08} & 4.29 & \spd{1.67} & 4.02 & \spd{1.56} & 3.95 & \spd{1.68} & 3.86 & \spd{1.58} \\
      & DFlash-z~\textcolor{gray}{\scriptsize(16)} & 4.92 & \spd{2.76} & 5.63 & \spd{3.10} & 3.81 & \spd{2.35} & 5.01 & \spd{2.62} & 4.82 & \spd{2.56} & 4.43 & \spd{2.45} & 4.34 & \spd{2.50} & 4.71 & \spd{2.62} \\
      & DFlash~\textcolor{gray}{\scriptsize(16)}  & 5.43 & \spd{2.97} & 5.60 & \spd{3.13} & 4.05 & \spd{2.31} & 5.17 & \spd{2.93} & 5.57 & \spd{2.57} & 5.16 & \spd{2.53} & 4.84 & \spd{2.62} & 5.12 & \spd{2.72} \\
      & MMFlash~\textcolor{gray}{\scriptsize(16)} & \textbf{5.97} & \textbf{\spd{3.28}} & \textbf{6.17} & \textbf{\spd{3.46}} & \textbf{4.49} & \textbf{\spd{2.55}} & \textbf{5.68} & \textbf{\spd{3.19}} & \textbf{6.26} & \textbf{\spd{2.87}} & \textbf{6.00} & \textbf{\spd{2.78}} & \textbf{5.35} & \textbf{\spd{2.88}} & \textbf{5.70} & \textbf{\spd{3.00}} \\
    \midrule
    \multirow{7}{*}{Qwen3.5-9B}
      & MTP~\textcolor{gray}{\scriptsize(3)}  & 3.32 & \spd{1.83} & 3.28 & \spd{1.87} & 3.01 & \spd{1.73} & 2.88 & \spd{1.57} & 3.38 & \spd{1.80} & 3.26 & \spd{1.80} & 3.24 & \spd{1.80} & 3.20 & \spd{1.77} \\
      & MTP~\textcolor{gray}{\scriptsize(7)}  & 4.66 & \spd{1.88} & 4.60 & \spd{1.94} & 3.91 & \spd{1.66} & 3.55 & \spd{1.41} & 4.74 & \spd{1.77} & 4.52 & \spd{1.78} & 4.43 & \spd{1.77} & 4.34 & \spd{1.74} \\
      & MTP~\textcolor{gray}{\scriptsize(15)} & 5.27 & \spd{1.38} & 5.27 & \spd{1.47} & 4.20 & \spd{1.18} & 3.75 & \spd{0.99} & 5.25 & \spd{1.30} & 5.08 & \spd{1.32} & 4.90 & \spd{1.30} & 4.81 & \spd{1.27} \\
      & EAGLE-3~\textcolor{gray}{\scriptsize(16)}  & 3.93 & \spd{1.81} & 3.85 & \spd{1.93} & 3.06 & \spd{1.55} & 2.72 & \spd{1.32} & 3.60 & \spd{1.76} & 3.50 & \spd{1.63} & 3.48 & \spd{1.75} & 3.45 & \spd{1.68} \\
      & DFlash-z~\textcolor{gray}{\scriptsize(16)} & 5.22 & \spd{3.03} & 5.27 & \spd{3.15} & 3.93 & \spd{2.46} & 3.67 & \spd{2.11} & 4.96 & \spd{2.75} & 4.77 & \spd{2.75} & 4.52 & \spd{2.67} & 4.62 & \spd{2.70} \\
      & DFlash~\textcolor{gray}{\scriptsize(16)}  & 5.59 & \spd{3.20} & 5.17 & \spd{3.09} & 4.15 & \spd{2.46} & 3.79 & \spd{2.20} & 5.62 & \spd{2.71} & 5.38 & \spd{2.83} & 4.92 & \spd{2.79} & 4.95 & \spd{2.75} \\
      & MMFlash~\textcolor{gray}{\scriptsize(16)} & \textbf{6.19} & \textbf{\spd{3.58}} & \textbf{5.67} & \textbf{\spd{3.42}} & \textbf{4.58} & \textbf{\spd{2.67}} & \textbf{4.20} & \textbf{\spd{2.39}} & \textbf{6.44} & \textbf{\spd{3.02}} & \textbf{6.49} & \textbf{\spd{3.17}} & \textbf{5.60} & \textbf{\spd{3.10}} & \textbf{5.60} & \textbf{\spd{3.05}} \\
    \midrule
    \multicolumn{2}{c}{Temperature $= 1$}
      & {\scriptsize $\tau$} & {\scriptsize Speedup} & {\scriptsize $\tau$} & {\scriptsize Speedup} & {\scriptsize $\tau$} & {\scriptsize Speedup}
      & {\scriptsize $\tau$} & {\scriptsize Speedup} & {\scriptsize $\tau$} & {\scriptsize Speedup} & {\scriptsize $\tau$} & {\scriptsize Speedup}
      & {\scriptsize $\tau$} & {\scriptsize Speedup}
      & {\scriptsize $\tau$} & {\scriptsize Speedup} \\
    \midrule
    \multirow{7}{*}{Qwen3.5-4B}
      & MTP~\textcolor{gray}{\scriptsize(3)}  & 3.04 & \spd{1.58} & 3.07 & \spd{1.60} & 2.74 & \spd{1.48} & 2.41 & \spd{1.34} & 3.17 & \spd{1.59} & 3.00 & \spd{1.54} & 3.00 & \spd{1.57} & 2.92 & \spd{1.53} \\
      & MTP~\textcolor{gray}{\scriptsize(7)}  & 3.93 & \spd{1.53} & 4.26 & \spd{1.68} & 3.41 & \spd{1.40} & 2.58 & \spd{1.14} & 4.22 & \spd{1.55} & 3.97 & \spd{1.49} & 3.91 & \spd{1.52} & 3.75 & \spd{1.48} \\
      & MTP~\textcolor{gray}{\scriptsize(15)} & 4.33 & \spd{1.14} & 4.76 & \spd{1.24} & 3.57 & \spd{0.99} & 2.62 & \spd{0.86} & 4.55 & \spd{1.14} & 4.33 & \spd{1.08} & 4.22 & \spd{1.11} & 4.06 & \spd{1.08} \\
      & EAGLE-3~\textcolor{gray}{\scriptsize(16)}  & 3.97 & \spd{1.59} & 3.97 & \spd{1.60} & 3.23 & \spd{1.38} & 2.60 & \spd{1.01} & 4.09 & \spd{1.49} & 3.87 & \spd{1.45} & 3.68 & \spd{1.53} & 3.63 & \spd{1.44} \\
      & DFlash-z~\textcolor{gray}{\scriptsize(16)} & 4.49 & \spd{2.42} & 5.00 & \spd{2.66} & 3.43 & \spd{2.06} & 4.81 & \spd{2.12} & 4.64 & \spd{2.41} & 4.14 & \spd{2.19} & 4.01 & \spd{2.28} & 4.36 & \spd{2.30} \\
      & DFlash~\textcolor{gray}{\scriptsize(16)}  & 4.87 & \spd{2.56} & 4.95 & \spd{2.66} & 3.71 & \spd{2.04} & 4.92 & \spd{2.48} & 5.07 & \spd{2.31} & 4.66 & \spd{2.22} & 4.47 & \spd{2.37} & 4.67 & \spd{2.38} \\
      & MMFlash~\textcolor{gray}{\scriptsize(16)} & \textbf{5.23} & \textbf{\spd{2.74}} & \textbf{5.39} & \textbf{\spd{2.86}} & \textbf{3.96} & \textbf{\spd{2.18}} & \textbf{5.25} & \textbf{\spd{2.58}} & \textbf{5.75} & \textbf{\spd{2.56}} & \textbf{5.23} & \textbf{\spd{2.37}} & \textbf{4.69} & \textbf{\spd{2.49}} & \textbf{5.07} & \textbf{\spd{2.54}} \\
    \midrule
    \multirow{7}{*}{Qwen3.5-9B}
      & MTP~\textcolor{gray}{\scriptsize(3)}  & 3.19 & \spd{1.72} & 3.08 & \spd{1.72} & 2.86 & \spd{1.61} & 2.78 & \spd{1.49} & 3.25 & \spd{1.70} & 3.11 & \spd{1.69} & 3.11 & \spd{1.71} & 3.05 & \spd{1.66} \\
      & MTP~\textcolor{gray}{\scriptsize(7)}  & 4.35 & \spd{1.72} & 4.15 & \spd{1.71} & 3.60 & \spd{1.49} & 3.37 & \spd{1.34} & 4.46 & \spd{1.64} & 4.26 & \spd{1.65} & 4.17 & \spd{1.66} & 4.05 & \spd{1.60} \\
      & MTP~\textcolor{gray}{\scriptsize(15)} & 4.84 & \spd{1.24} & 4.57 & \spd{1.27} & 3.81 & \spd{1.06} & 3.53 & \spd{0.93} & 4.92 & \spd{1.22} & 4.65 & \spd{1.18} & 4.56 & \spd{1.21} & 4.41 & \spd{1.16} \\
      & EAGLE-3~\textcolor{gray}{\scriptsize(16)}  & 3.60 & \spd{1.67} & 3.56 & \spd{1.72} & 2.84 & \spd{1.42} & 2.59 & \spd{1.23} & 3.46 & \spd{1.61} & 3.32 & \spd{1.50} & 3.29 & \spd{1.62} & 3.24 & \spd{1.54} \\
      & DFlash-z~\textcolor{gray}{\scriptsize(16)} & 4.77 & \spd{2.64} & 4.70 & \spd{2.76} & 3.66 & \spd{2.25} & 3.42 & \spd{1.97} & 4.71 & \spd{2.57} & 4.49 & \spd{2.51} & 4.26 & \spd{2.49} & 4.29 & \spd{2.46} \\
      & DFlash~\textcolor{gray}{\scriptsize(16)}  & 4.94 & \spd{2.75} & 4.50 & \spd{2.65} & 3.74 & \spd{2.18} & 3.55 & \spd{2.02} & 5.24 & \spd{2.52} & 4.85 & \spd{2.52} & 4.53 & \spd{2.57} & 4.48 & \spd{2.46} \\
      & MMFlash~\textcolor{gray}{\scriptsize(16)} & \textbf{5.54} & \textbf{\spd{3.22}} & \textbf{4.85} & \textbf{\spd{2.85}} & \textbf{4.13} & \textbf{\spd{2.37}} & \textbf{3.82} & \textbf{\spd{2.17}} & \textbf{6.07} & \textbf{\spd{2.77}} & \textbf{5.71} & \textbf{\spd{2.74}} & \textbf{5.10} & \textbf{\spd{2.81}} & \textbf{5.03} & \textbf{\spd{2.70}} \\
    \bottomrule
  \end{tabular*}
  \vspace{-3mm}
  \caption{Decoding speedup over baseline and average acceptance length ($\tau$) on Qwen3.5 models across various benchmarks. Parenthesized values indicate the number of draft steps for MTP, the draft tree size for EAGLE-3, and the diffusion block size for DFlash \& MMFlash.}
  \label{tab:speedup_qwen3}
  \vspace{-4mm}
\end{table*}

\noindent \textbf{Sparse Attention in MMFlash Drafter for Video-Length Contexts.}
We further examine video benchmarks, where the average input length increases from 477 tokens across seven image benchmarks to 41,218 tokens. Diffusion drafters such as DFlash attend to the entire target context at every drafting step, raising the question of whether full-context attention is necessary for a small drafter. We therefore analyze the attention patterns of the trained MMFlash drafter on video inputs. Figure~\ref{fig:sparse_motivation}(a) shows that attention concentrates on the question and generated answer, while visual tokens receive diffuse attention with no frame standing out. Across 60 questions, Figure~\ref{fig:sparse_motivation}(b) confirms this locality: the most recent 256 positions account for only 0.8\% of the context but receive 50.1\% of the attention. In contrast, distant frames occupy over 96\% of the context and collectively absorb 46\% of the attention despite low attention per token, suggesting substantial redundancy in full-context drafting. Furthermore, we apply a simple attention window to the trained MMFlash drafter. As shown in Figure~\ref{fig:sparse_motivation}(c), windowing reduces acceptance length on image benchmarks but improves it on video benchmarks, supporting the benefit of sparse attention in long video contexts. A plain window discards visual evidence outside its range, potentially removing information needed to answer the question. Based on the observed attention patterns, we design a simple sparse attention pattern specifically for video benchmarks. It retains the first 4 sink tokens, visual tokens sampled with a stride of 32, all text tokens, and a local window of 256 tokens. We use the same pattern during training and inference to ensure consistent context access. Implementation details are provided in Section~\ref{subsec:main-results}.

\section{Experiments}
\label{sec:experiments}

\subsection{Experimental Setup}
\label{sec:setup}

\begin{table}[t]
  \centering
  \small
  \setlength{\tabcolsep}{1.5pt}
  \renewcommand{\arraystretch}{0.95}
  \begin{tabular*}{\linewidth}{@{\extracolsep{\fill}}l*{6}{c}@{}}
    \toprule
    \multirow{3}{*}{Method}
      & \multicolumn{2}{c}{LongVideoBench} & \multicolumn{2}{c}{MVBench} & \multicolumn{2}{c}{Video-MME} \\
      & \multicolumn{2}{c}{\textcolor{gray}{\scriptsize $\sim$\,41.6\,K tokens}}
      & \multicolumn{2}{c}{\textcolor{gray}{\scriptsize $\sim$\,42.3\,K tokens}}
      & \multicolumn{2}{c}{\textcolor{gray}{\scriptsize $\sim$\,39.8\,K tokens}} \\
    \cmidrule(lr){2-3} \cmidrule(lr){4-5} \cmidrule(lr){6-7}
      & {\scriptsize $\tau$} & {\scriptsize Speedup} & {\scriptsize $\tau$} & {\scriptsize Speedup}
      & {\scriptsize $\tau$} & {\scriptsize Speedup} \\
    \midrule
    \multicolumn{7}{@{}l}{\textbf{Qwen3.5-4B}} \\
    \midrule
    MTP~\textcolor{gray}{\scriptsize(15)}      & 3.99 & \spd{1.31} & 3.80 & \spd{1.09} & \textbf{3.50} & \spd{1.00} \\
    EAGLE-3~\textcolor{gray}{\scriptsize(4)}   & 2.04 & \spd{1.63} & 1.91 & \spd{1.45} & 1.92 & \spd{1.43} \\
    DFlash~\textcolor{gray}{\scriptsize(16)}   & 3.14 & \spd{2.11} & 3.01 & \spd{1.82} & 2.81 & \spd{1.75} \\
    MMFlash~\textcolor{gray}{\scriptsize(16)}  & 3.23 & \spd{2.23} & 2.91 & \spd{1.83} & 2.86 & \spd{1.82} \\
    \quad + sparse attn.                       & 3.81 & \spd{2.44} & 3.50 & \spd{2.05} & 3.20 & \spd{1.93} \\
    \quad + fine-tuning                        & \textbf{4.09} & \textbf{\spd{2.67}} & \textbf{3.83} & \textbf{\spd{2.24}} & 3.45 & \textbf{\spd{2.09}} \\
    \midrule
    \multicolumn{7}{@{}l}{\textbf{Qwen3.5-9B}} \\
    \midrule
    MTP~\textcolor{gray}{\scriptsize(15)}      & 4.07 & \spd{1.34} & \textbf{3.94} & \spd{1.26} & \textbf{3.73} & \spd{1.10} \\
    EAGLE-3~\textcolor{gray}{\scriptsize(4)}   & 1.88 & \spd{1.29} & 1.67 & \spd{1.06} & 1.79 & \spd{1.25} \\
    DFlash~\textcolor{gray}{\scriptsize(16)}   & 3.56 & \spd{2.71} & 3.19 & \spd{2.37} & 3.21 & \spd{2.22} \\
    MMFlash~\textcolor{gray}{\scriptsize(16)}  & 3.63 & \spd{2.79} & 3.23 & \spd{2.42} & 3.28 & \spd{2.27} \\
    \quad + sparse attn.                       & 3.81 & \spd{2.81} & 3.46 & \spd{2.49} & 3.42 & \spd{2.21} \\
    \quad + fine-tuning                        & \textbf{4.16} & \textbf{\spd{3.07}} & 3.74 & \textbf{\spd{2.66}} & 3.63 & \textbf{\spd{2.39}} \\
    \bottomrule
  \end{tabular*}
  \vspace{-3mm}
  \caption{Average acceptance length ($\tau$) and decoding speedup over baseline on three video benchmarks at temperature $= 0$. The gray number under each benchmark gives its average context length.} 
  \label{tab:video}
  \vspace{-4mm}
\end{table}

We summarize the experimental setup below and provide more details in Appendix~\ref{supp:experimental_setup}.

\noindent \textbf{Training Data.}
To ensure data diversity, we collect 1M samples from LLaVA-OV-1.5-Instruct-Data and regenerate their responses using the target model for better alignment. We modify selected prompts to elicit longer responses for training efficiency (Appendix~\ref{supp:training_data}). We follow the official generation settings, with temperature set to 0.0, top-$p$ to 0.95, top-$k$ to 20, and a maximum generation length of 4,096 tokens. After discarding samples whose total sequence length exceeds 4,096 tokens, we obtain approximately 900K training samples.

\noindent \textbf{Training Details.}
We adopt Qwen3.5-4B and Qwen3.5-9B~\citep{qwen3.5}, two popular open-source VLMs, as our target models. For training, we mainly follow the official setting of DFlash. Specifically, we set the block size to 16 and the number of anchors to 512, use a batch size of 2, and train the drafter for 6 epochs. The learning rate is set to $6.0 \times 10^{-4}$ and follows a cosine annealing schedule that decays to zero, preceded by a linear warmup over the first 4\% of training steps. All training is conducted on 8 NVIDIA H100 GPUs.

\noindent \textbf{Baselines.}
We compare MMFlash against the following baselines: (1) \textbf{MTP}, the multi-token prediction module natively supported by the Qwen3.5 series, for which we report results under different numbers of draft steps; (2) \textbf{EAGLE-3}, the state-of-the-art speculative decoding method~\citep{li2026eagle}, which we train on the same data as MMFlash; (3) \textbf{DFlash-z}, the officially released DFlash drafter, which consists of 6 layers, extracts hidden states from 8 layers of the target model, and is trained on text-only data\footnote{The full training recipe of this drafter has not been publicly released.}; and (4) \textbf{DFlash}, which we train on the same data as MMFlash following the official setting, \ie, a 5-layer drafter that extracts hidden states from 5 target layers. For a fair comparison, all trained baselines share the training data of MMFlash, and all other unrelated hyperparameters are kept identical.

\noindent \textbf{Inference Details.}
We follow the official inference settings released with each target model, with thinking mode disabled and a maximum of 4,096 generated tokens. For all baselines and MMFlash, we set top-$p$ to 0.95 and top-$k$ to 20, and evaluate under two temperatures, $T=0$ and $T=1$. For the main experiments, the block size of the drafter is set to 16, consistent with training, and the concurrency is fixed to 1. All inference experiments are run on a single NVIDIA H200 GPU with SGLang~\citep{zheng2024sglang} framework.

\noindent \textbf{Evaluations.} 
We evaluate our method on seven multimodal benchmarks covering text recognition and document understanding (ChartQA~\citep{masry2022chartqa} and CharXiv~\citep{wang2024charxiv}), general visual question answering (MMStar~\citep{chen2024we} and MMBench~\citep{liu2024mmbench}), and STEM tasks (DynaMath~\citep{zou2025dynamath}, MathVista~\citep{lu2024mathvista}, and MathVerse~\citep{zhang2024mathverse}). All evaluations are conducted in a zero-shot setting on the same 200 random samples per benchmark (Appendix~\ref{supp:inference_eval}), following the LMMs-Eval framework\footnote{\url{https://github.com/evolvinglmms-lab/lmms-eval}}.

\subsection{Main Results}
\label{subsec:main-results}

\noindent \textbf{MMFlash Achieves the Highest Acceptance Length and Speedup.}
As shown in Table~\ref{tab:speedup_qwen3}, MMFlash consistently outperforms all baselines in both acceptance length and speedup across benchmarks, target models, and temperatures. It achieves average decoding speedups of 3.00-3.05$\times$ under greedy decoding ($T=0$) and 2.54-2.70$\times$ at $T=1$ across the two target models, 6.7-10.9\% higher than DFlash. Overall, these results demonstrate the effectiveness and robustness of MMFlash.

\begin{table}[t]
  \centering
  \small
  \setlength{\tabcolsep}{1.5pt}
  \renewcommand{\arraystretch}{0.95}
  \begin{tabular*}{\linewidth}{@{\extracolsep{\fill}}ll*{5}{c}c@{}}
    \toprule
    \multirow{2}{*}{Task} & \multirow{2}{*}{Method} & \multicolumn{5}{c}{Concurrency} & \multirow{2}{*}{\textit{Avg.} $\tau$} \\
    \cmidrule(lr){3-7}
    & & 1 & 4 & 8 & 16 & 32 & \\
    \midrule
    \multicolumn{8}{@{}l}{\textbf{Qwen3.5-4B}} \\
    \midrule
    \multirow{3}{*}{CharXiv}
      & Baseline & 238 & 794 & 1306 & 2094 & 2733 & -- \\
    \cmidrule(l){2-8}
      & \multirow{2}{*}{MMFlash} & 826 & 2054 & 3102 & 3943 & 4395 & \multirow{2}{*}{6.23} \\
      &                          & \spd{3.46} & \spd{2.59} & \spd{2.37} & \spd{1.88} & \spd{1.61} & \\
    \midrule
    \multirow{3}{*}{MMBench}
      & Baseline & 242 & 453 & 672 & 1141 & 1748 & -- \\
    \cmidrule(l){2-8}
      & \multirow{2}{*}{MMFlash} & 772 & 1032 & 1451 & 1929 & 2364 & \multirow{2}{*}{5.69} \\
      &                          & \spd{3.19} & \spd{2.28} & \spd{2.16} & \spd{1.69} & \spd{1.35} & \\
    \midrule
    \multirow{3}{*}{DynaMath}
      & Baseline & 239 & 836 & 1456 & 2276 & 3440 & -- \\
    \cmidrule(l){2-8}
      & \multirow{2}{*}{MMFlash} & 685 & 1991 & 3081 & 4066 & 4643 & \multirow{2}{*}{6.27} \\
      &                          & \spd{2.87} & \spd{2.38} & \spd{2.12} & \spd{1.79} & \spd{1.35} & \\
    \midrule
    \multicolumn{8}{@{}l}{\textbf{Qwen3.5-9B}} \\
    \midrule
    \multirow{3}{*}{ChartQA}
      & Baseline & 174 & 633 & 1048 & 1694 & 1952 & -- \\
    \cmidrule(l){2-8}
      & \multirow{2}{*}{MMFlash} & 623 & 1732 & 2438 & 3080 & 3685 & \multirow{2}{*}{6.21} \\
      &                          & \spd{3.58} & \spd{2.73} & \spd{2.33} & \spd{1.82} & \spd{1.89} & \\
    \midrule
    \multirow{3}{*}{MMStar}
      & Baseline & 172 & 633 & 1064 & 1697 & 2462 & -- \\
    \cmidrule(l){2-8}
      & \multirow{2}{*}{MMFlash} & 460 & 1380 & 1964 & 2511 & 2814 & \multirow{2}{*}{4.59} \\
      &                          & \spd{2.67} & \spd{2.18} & \spd{1.85} & \spd{1.48} & \spd{1.14} & \\
    \midrule
    \multirow{3}{*}{DynaMath}
      & Baseline & 172 & 655 & 1155 & 1884 & 2665 & -- \\
    \cmidrule(l){2-8}
      & \multirow{2}{*}{MMFlash} & 520 & 1645 & 2541 & 3308 & 3792 & \multirow{2}{*}{6.49} \\
      &                          & \spd{3.02} & \spd{2.51} & \spd{2.20} & \spd{1.76} & \spd{1.42} & \\
    \bottomrule
  \end{tabular*}
  \vspace{-3mm}
  \caption{Decode throughput (tok/s), speedup over the baseline, and average acceptance length ($\tau$) on SGLang under increasing serving concurrency. For each MMFlash block the upper line reports throughput and the lower line the speedup at the same concurrency.}
  \label{tab:concurrency_sglang}
  \vspace{-4mm}
\end{table}

\begin{table*}[t]
  \centering
  \small
  \setlength{\tabcolsep}{2pt}
  \begin{tabular*}{\linewidth}{@{\extracolsep{\fill}}ll*{16}{c}@{}}
    \toprule
    \multirow{3}{*}{Model} & \multirow{3}{*}{Method}
      & \multicolumn{4}{c}{\textsc{Text \& Doc}} & \multicolumn{4}{c}{\textsc{General VQA}} & \multicolumn{6}{c}{\textsc{STEM}} & \multicolumn{2}{c}{\multirow{2}{*}{\textit{Avg.}}} \\
    \cmidrule(lr){3-6} \cmidrule(lr){7-10} \cmidrule(lr){11-16}
    & & \multicolumn{2}{c}{ChartQA} & \multicolumn{2}{c}{CharXiv}
      & \multicolumn{2}{c}{MMStar} & \multicolumn{2}{c}{MMBench}
      & \multicolumn{2}{c}{DynaMath} & \multicolumn{2}{c}{MathVista} & \multicolumn{2}{c}{MathVerse}
      & \multicolumn{2}{c}{} \\
    & & {\scriptsize $\tau$} & {\scriptsize Speedup} & {\scriptsize $\tau$} & {\scriptsize Speedup} & {\scriptsize $\tau$} & {\scriptsize Speedup}
      & {\scriptsize $\tau$} & {\scriptsize Speedup} & {\scriptsize $\tau$} & {\scriptsize Speedup} & {\scriptsize $\tau$} & {\scriptsize Speedup}
      & {\scriptsize $\tau$} & {\scriptsize Speedup}
      & {\scriptsize $\tau$} & {\scriptsize Speedup} \\
    \midrule
    \multirow{2}{*}{Qwen3.5-4B}
      & MMFlash~\textcolor{gray}{\scriptsize(8)}  & 4.70 & \spd{3.01} & 4.77 & \spd{3.13} & 3.80 & \spd{2.49} & 4.52 & \spd{2.89} & 4.92 & \spd{2.80} & 4.68 & \spd{2.74} & 4.43 & \spd{2.83} & 4.55 & \spd{2.84} \\
      & MMFlash~\textcolor{gray}{\scriptsize(16)} & 5.97 & \spd{3.28} & 6.17 & \spd{3.46} & 4.49 & \spd{2.55} & 5.68 & \spd{3.19} & 6.26 & \spd{2.87} & 6.00 & \spd{2.78} & 5.35 & \spd{2.88} & 5.70 & \spd{3.00} \\
    \midrule
    \multirow{2}{*}{Qwen3.5-9B}
      & MMFlash~\textcolor{gray}{\scriptsize(8)}   & 5.07 & \spd{3.22} & 4.72 & \spd{3.17} & 4.01 & \spd{2.65} & 3.77 & \spd{2.30} & 5.18 & \spd{2.94} & 5.02 & \spd{2.96} & 4.68 & \spd{2.95} & 4.64 & \spd{2.89} \\
      & MMFlash~\textcolor{gray}{\scriptsize(16)}  & 6.19 & \spd{3.58} & 5.67 & \spd{3.42} & 4.58 & \spd{2.67} & 4.20 & \spd{2.39} & 6.44 & \spd{3.02} & 6.49 & \spd{3.17} & 5.60 & \spd{3.10} & 5.60 & \spd{3.05} \\
    \bottomrule
  \end{tabular*}
  \vspace{-3mm}
  \caption{Effect of the test-time block size on MMFlash. Parenthesized values indicate the test-time block size for MMFlash. All drafters are trained with block size 16.}
  \label{tab:block_size}
  \vspace{-4mm}
\end{table*}

\begin{table}[t]
  \centering
  \small
  \setlength{\tabcolsep}{1.5pt}
  \renewcommand{\arraystretch}{0.95}
  \begin{tabular*}{\linewidth}{@{\extracolsep{\fill}}l*{6}{c}@{}}
    \toprule
    \multirow{2}{*}{Method}
      & \multicolumn{2}{c}{ChartQA} & \multicolumn{2}{c}{MMStar} & \multicolumn{2}{c}{DynaMath} \\
    \cmidrule(lr){2-3} \cmidrule(lr){4-5} \cmidrule(lr){6-7}
      & {\scriptsize $\tau$} & {\scriptsize Speedup} & {\scriptsize $\tau$} & {\scriptsize Speedup}
      & {\scriptsize $\tau$} & {\scriptsize Speedup} \\
    \midrule
    \multicolumn{7}{@{}l}{\textbf{Gemma4-E4B-it}} \\
    \midrule
    MTP~\textcolor{gray}{\scriptsize(5, official)}   & 3.81 & \spd{1.78} & 3.19 & \spd{1.48} & 4.04 & \spd{1.89} \\
    MTP~\textcolor{gray}{\scriptsize(15)}  & 4.73 & \spd{1.66} & 3.74 & \spd{1.33} & 5.17 & \spd{1.75} \\
    EAGLE-3~\textcolor{gray}{\scriptsize(16)}  & 4.87 & \spd{2.27} & 4.04 & \spd{1.50} & 6.23 & \spd{1.90} \\
    MMFlash~\textcolor{gray}{\scriptsize(16)}  & \textbf{6.39} & \textbf{\spd{2.87}} & \textbf{4.83} & \textbf{\spd{2.10}} & \textbf{7.15} & \textbf{\spd{3.02}} \\
    \midrule
    \multicolumn{7}{@{}l}{\textbf{Qwen3.8-27B}} \\
    \midrule
    MTP~\textcolor{gray}{\scriptsize(7)}       & 5.19 & \spd{2.90} & 4.55 & \spd{2.59} & 5.48 & \spd{3.07} \\
    MTP~\textcolor{gray}{\scriptsize(15)}      & 6.36 & \spd{2.47} & 5.25 & \spd{2.16} & 6.63 & \spd{2.68} \\
    MMFlash~\textcolor{gray}{\scriptsize(16)}  & \textbf{6.37} & \textbf{\spd{4.18}} & \textbf{5.36} & \textbf{\spd{3.54}} & \textbf{7.57} & \textbf{\spd{4.71}} \\
    \bottomrule
  \end{tabular*}
  \vspace{-3mm}
  \caption{Average acceptance length ($\tau$) and decoding speedup on Gemma4-E4B-it and Qwen3.8-27B at temperature $= 0$.} %
  \label{tab:generalization}
  \vspace{-3mm}
\end{table}

\begin{table}[t]
  \centering
  \small
  \setlength{\tabcolsep}{1.5pt}
  \renewcommand{\arraystretch}{0.95}
  \begin{tabular*}{\linewidth}{@{\extracolsep{\fill}}l*{6}{c}@{}}
    \toprule
    \multirow{2}{*}{Setting}
      & \multicolumn{2}{c}{CharXiv} & \multicolumn{2}{c}{MMBench} & \multicolumn{2}{c}{DynaMath} \\
    \cmidrule(lr){2-3} \cmidrule(lr){4-5} \cmidrule(lr){6-7}
      & {\scriptsize $\tau$} & {\scriptsize Speedup} & {\scriptsize $\tau$} & {\scriptsize Speedup}
      & {\scriptsize $\tau$} & {\scriptsize Speedup} \\
    \midrule
    3-L & 5.56 & \spd{3.25} & 4.69 & \spd{2.86} & 5.78 & \spd{2.70} \\
    5-L & 6.17 & \spd{3.46} & 5.68 & \spd{3.19} & 6.26 & \spd{2.87} \\
    8-L & 6.39 & \spd{3.42} & 6.21 & \spd{3.21} & 6.69 & \spd{2.88} \\
    \bottomrule
  \end{tabular*}
  \vspace{-3mm}
  \caption{Effect of the number of drafter layers on MMFlash for Qwen3.5-4B at temperature $= 0$. $n$-L denotes a drafter with $n$ layers; all drafters are trained with block size 16 and read hidden features from 5 layers of the target model.}
  \label{tab:drafter_scale}
  \vspace{-4mm}
\end{table}

\noindent \textbf{MMFlash Remains Superior on Video Benchmarks.}
We further evaluate MMFlash on three video benchmarks: LongVideoBench~\citep{wu2024longvideobench}, MVBench~\citep{li2024mvbench}, and Video-MME~\citep{fu2025video}. We fine-tune the MMFlash drafter for 3 epochs on 2K samples from LLaVA-Video-178K~\citep{zhang2024videoinstruction} using the proposed sparse attention pattern. As shown in Table~\ref{tab:video}, sparse attention alone, without any training, improves acceptance length by 11.1\% and speedup by 4.7\% on average. Fine-tuning increases these gains to 20.0\% and 13.7\%, respectively, yielding average speedups of 2.33-2.71$\times$, 11.2-23.2\% higher than DFlash, at minimal fine-tuning cost.

\noindent \textbf{MMFlash Generalizes Across Model Families and Scales.}
We assess MMFlash's generalization across model families and scales using Gemma4-E4B-it~\citep{team2026gemma} and Qwen3.8-27B~\citep{qwen38}. We train the EAGLE-3 and MMFlash drafters on the same data regenerated by the target model. As shown in Table~\ref{tab:generalization}, MMFlash achieves the highest acceptance length and decoding speedup on all three benchmarks for both models, with speedups of 2.10-3.02$\times$ on Gemma and 3.54-4.71$\times$ on Qwen3.8-27B.

\subsection{Diagnostic Results}

In this section, we first evaluate MMFlash under the high serving concurrency of real-world deployment, and then ablate three key design choices of the MMFlash drafter: the test-time block size, the drafter scale, and the number of target hidden states. All ablated drafters are trained on the same training data described in Section~\ref{sec:setup}.

\noindent \textbf{MMFlash Remains Effective under High Serving Concurrency.}
As shown in Table~\ref{tab:concurrency_sglang}, MMFlash accelerates decoding at all concurrency levels from 1 to 32, peaking at 3.58$\times$ on Qwen3.5-9B. Although the speedup narrows as decoding becomes compute-bound, MMFlash at concurrency 16 still delivers higher throughput than the baseline at concurrency 32 on all six tasks, confirming its effectiveness in real-world serving. Results on vLLM~\citep{kwon2023efficient} are consistent (as shown in Table~\ref{tab:concurrency_vllm} in Appendix~\ref{supp:vllm}).

\noindent \textbf{Ablation on Test-Time Block Size.} 
Since the block size is a core design choice of MMFlash, we first ablate the test-time block size by evaluating MMFlash with block sizes of 8 and 16, where 16 matches the block length used during training. As shown in Table~\ref{tab:block_size}, MMFlash performs best with a block size of 16, which is aligned with the training configuration. When the test-time block size is smaller than that used in training, MMFlash remains robust: the average speedup decreases by only 5.3\% and still surpasses all baselines in Table~\ref{tab:speedup_qwen3}. In contrast, the average acceptance length drops by 18.7\%, since a smaller block caps the number of tokens accepted per draft block and thus under-utilizes the drafter. For instance, 28.8\% and 27.6\% of draft blocks saturate the maximum acceptance length of 8 on ChartQA and CharXiv, respectively. Overall, the MMFlash drafter generalizes well to smaller test-time block sizes, which supports dynamic block-size scheduling in real-world serving.

\noindent \textbf{Ablation on the Scale of MMFlash Drafter.}
As shown in Table~\ref{tab:drafter_scale}, increasing the depth of the MMFlash drafter from 3 to 8 layers improves the acceptance length, and the same trend holds at temperature $= 1$ (Table~\ref{tab:drafter_scale_t1} in Appendix~\ref{supp:additional_results}). This indicates that the capability of the MMFlash drafter scales with its size. 
However, a deeper drafter also costs more per drafting pass: the average speedup rises by 6.7-8.1\% from 3 to 5 layers but changes by less than 1\% from 5 to 8 layers at both temperatures, so we default to 5 layers. Overall, the drafter scale should be chosen according to the application scenario to achieve the best trade-off between draft quality and drafting cost.

\begin{table}[t]
  \centering
  \small
  \setlength{\tabcolsep}{1.5pt}
  \renewcommand{\arraystretch}{0.95}
  \begin{tabular*}{\linewidth}{@{\extracolsep{\fill}}l*{6}{c}@{}}
    \toprule
    \multirow{2}{*}{Setting}
      & \multicolumn{2}{c}{CharXiv} & \multicolumn{2}{c}{MMBench} & \multicolumn{2}{c}{DynaMath} \\
    \cmidrule(lr){2-3} \cmidrule(lr){4-5} \cmidrule(lr){6-7}
      & {\scriptsize $\tau$} & {\scriptsize Speedup} & {\scriptsize $\tau$} & {\scriptsize Speedup}
      & {\scriptsize $\tau$} & {\scriptsize Speedup} \\
    \midrule
    5-H & 6.17 & \spd{3.46} & 5.68 & \spd{3.19} & 6.26 & \spd{2.87} \\
    8-H & 6.18 & \spd{3.47} & 6.01 & \spd{3.20} & 6.37 & \spd{2.88} \\
    \bottomrule
  \end{tabular*}
  \vspace{-3mm}
  \caption{Effect of the number of tapped target hidden states on MMFlash for Qwen3.5-4B at temperature $= 0$. $n$-H denotes a drafter that reads hidden features from $n$ layers of the target model; all drafters have 5 layers and are trained with block size 16.}
  \label{tab:target_taps}
  \vspace{-4mm}
\end{table}

\noindent \textbf{Ablation on the Number of Target Hidden States.}
As shown in Table~\ref{tab:target_taps} (and Table~\ref{tab:target_taps_t1} in Appendix~\ref{supp:additional_results} for $T=1$), increasing the number of tapped target hidden states from 5 to 8 consistently improves acceptance length, suggesting that access to more layers provides richer semantic information. However, the speedup gain is marginal, at most $0.01\times$. Additional taps also increase training costs, requiring more storage for offline training and more time for online training (Appendix~\ref{supp:training_details}). In our setting, increasing the number of taps from 5 to 8 increases online training time by 19.9\%, so we use 5 taps by default.

\section{Conclusion}
\label{sec:conclusion}

We present MMFlash, a novel diffusion drafting framework for speculative decoding in VLMs. We find that draft acceptance drops on visually dependent tokens, even after multimodal training. MMFlash therefore adaptively upweights these tokens while the drafter still predicts them poorly, and adopts sparse attention for long videos. It achieves the best average decoding speedups across seven image and three video benchmarks, remains effective under high serving concurrency, and generalizes to other model families, offering a strong baseline for future work.

\section*{LLM Usage Statement}

We used Claude only to polish the writing of this paper, improving the grammar, wording, and readability of the text in all sections. All research ideas, methods, experiments, and results are by the authors, who reviewed all text and take full responsibility for the content of this paper.

{
    \small
    \bibliographystyle{ieeenat_fullname}
    \bibliography{main}
}

\clearpage
\setcounter{page}{1}
\maketitlesupplementary

\setcounter{section}{0}
\renewcommand{\thesection}{\Alph{section}}
\renewcommand{\theHsection}{supp.\Alph{section}} %

\section{Extended Related Work}
\label{supp:related}

\noindent \textbf{Speculative Decoding for VLMs.}
Speculative decoding~\citep{leviathan2023fast,cai2024medusa,li2024eagle1,chen2026dflash} accelerates autoregressive generation by having a lightweight draft model propose several candidate tokens that the target model verifies in a single parallel forward pass, leaving the target's output distribution unchanged. Work has since pursued drafters that are cheaper yet better aligned with the target, ranging from auxiliary decoding heads~\citep{cai2024medusa} to feature-level drafting with tree-structured verification~\citep{li2024eagle1,li2024eagle2,li2026eagle} and, most recently, drafters built on diffusion language models~\citep{chen2026dflash,liu2026dart}.

Bringing speculative decoding to VLMs introduces a design choice absent from the text-only setting: how the drafter should handle visual input. Existing methods fall into three paradigms according to this choice.
The first exposes the drafter to the full visual input, strengthening its visual perception so as to raise the acceptance rate~\citep{lin2025speculative,huo2025spec,ganesan2025massv,tong2026sage}. The second supplies only distilled visual information, preserving the essential multimodal semantics while substantially reducing the drafter's computation~\citep{hu2026dream,kang2026vispec,zhang2025sparse,ji2025specvlm,lv2026hippo,kong2026parallelvlm}. The third, which this work adopts, removes visual tokens from the drafter altogether~\citep{gagrani2024speculative,xie2026hivis,zhang2026sparrow}. Since the target model has already fused the visual content into its hidden states during prefill, the drafter need not re-encode the visual input; its cost therefore no longer scales with the number of visual tokens, making this the fastest of the three paradigms. 
Rather than committing to a single paradigm, several methods select the visual-input strategy adaptively at inference time according to the task at hand~\citep{huang2025specvlm,lee2026tabed,lee2024batch}. Orthogonal to the visual-input axis, other efforts relax the verification criterion to admit more draft tokens~\citep{ji2026see}, lift speculation from the token level to the agent level~\citep{huang2026speceyes}, or tailor speculative decoding to vertical domains such as document OCR~\citep{liao2026hsd}.

\noindent \textbf{Speculative Decoding with Diffusion Drafters.}
Diffusion large language models (dLLMs)~\citep{sahoo2024simple,zheng2024masked,nie2026large,ye2025dream} recast text generation as iterative denoising: starting from a fully masked sequence, the model predicts every masked position in parallel and, over a small number of refinement steps, commits to confident tokens while re-masking uncertain ones. Because generation is not bound to a left-to-right order, a dLLM can emit many tokens in a single forward pass, offering test-time flexibility and efficiency that autoregressive decoding lacks, and recent large-scale DLMs achieve quality competitive with autoregressive LLMs~\citep{nie2026large,khanna2025mercury,song2025seed}.
Block diffusion~\citep{arriola2025block} bridges the two families by running diffusion within fixed-size blocks that are themselves generated autoregressively, recovering KV caching and variable-length generation while retaining intra-block parallelism.

These properties make DLMs natural drafters: whereas an autoregressive drafter spends one forward pass per token, a DLM can propose a whole block at once.
Early explorations either blend the two paradigms within a single model or retrofit autoregressive models for parallel drafting. \citet{liu2511tidar} jointly trains diffusion and autoregressive objectives to ``think'' in parallel and ``talk'' sequentially, at the cost of exact losslessness, while \citet{samragh2025your} exploit the future-token information implicitly encoded in autoregressive LLMs, adding a LoRA adapter for parallel drafting and keeping the base model as verifier.
DiffuSpec~\citep{li2026diffuspec} and SpecDiff-2~\citep{sandler2026specdiff} instead draft directly with large pre-trained DLMs, relying on inference-time search or train--test alignment to raise acceptance.
DFlash~\citep{chen2026dflash} shows that the idea can be made simple and effective: a lightweight diffusion drafter emits an entire block of candidates in one forward pass, which the target then verifies as in standard speculative decoding.
Subsequent work refines this recipe: DART~\citep{liu2026dart} pursues a similar design; D-PACE~\citep{wu2026d} turns DFlash's fixed position-wise loss weights into a dynamic schedule; Domino~\citep{huang2026domino} adds causal dependencies within a block so that later positions see earlier committed tokens rather than only masks; and Nemotron~\citep{fu2026nemotron} unifies autoregressive, diffusion, and self-speculative decoding in a single tri-mode model that drafts with diffusion and verifies autoregressively.

\section{Block-Level Results of the Empirical Study}
\label{supp:empirical_study}

\begin{figure*}[t]
\centering
\includegraphics[width=\linewidth]{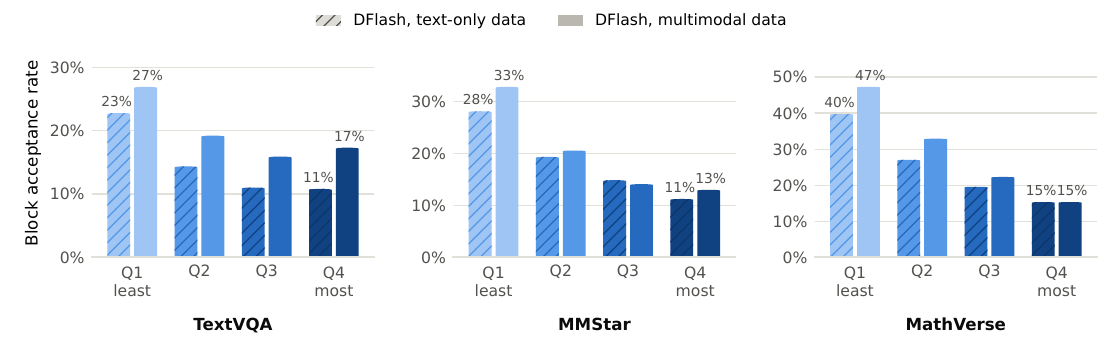}  \\
\vspace{-4mm}
\caption{Block acceptance rate on Qwen3.5-4B with three benchmarks, under the same setup as Figure~\ref{fig:motivation}. Each drafted block is grouped by the visual dependence quartile of its first predicted token. Visual dependence increases from Q1 to Q4.}
\label{fig:motivation_supp}
\vspace{-5mm}
\end{figure*}

Figure~\ref{fig:motivation_supp} extends the token-level study in Figure~\ref{fig:motivation} to drafted blocks, grouped by the visual-dependence quartile of their first predicted token. The trend is the same: averaged over benchmarks and drafters, block acceptance drops steadily from 33\% in Q1 to 14\% in Q4, a 2.4$\times$ gap. Training on multimodal data does not close this gap: it raises block acceptance by about 5 points in Q1 but by less than 3 points in Q4. These results further support the observation in Section~\ref{sec:motivation} that the acceptance loss of DFlash on VLMs concentrates on visually dependent tokens.

\section{Additional Details of MMFlash}
\label{supp:mmflash_details}

\subsection{Quantile Normalization of Visual Dependence}
\label{supp:quantile}

Raw KL values are unbounded and heavy-tailed, so we first map them to the unit interval by their rank in the corpus $\mathcal{D}$ of all scored tokens,
\begin{equation}
  \tilde g_t = F_{\mathcal{D}}(d_t)
             = \frac{1}{|\mathcal{D}|}\sum_{t'\in\mathcal{D}}\mathbf{1}\bigl[d_{t'}\le d_t\bigr],
  \label{eq:quantile}
\end{equation}
which is distribution-free: the quartiles of Figure~\ref{fig:motivation} correspond to $\tilde g_t\in[0,0.25)$, $[0.25,0.5)$, $[0.5,0.75)$ and $[0.75,1]$.

\section{Additional Experimental Setup}
\label{supp:experimental_setup}

\subsection{Training Data Construction}
\label{supp:training_data}

We regenerate all responses in the training set with the target model (greedy decoding, thinking disabled, at most 4{,}096 new tokens), and before regeneration we rewrite part of the prompts so that the target produces longer responses, which makes training more efficient. During training, the draft model is supervised on blocks that start at anchor positions sampled from the response tokens, with up to 512 anchors per sample. A short response therefore yields only a few training blocks, even though its image and prompt still cost a full forward pass of the target model, and samples whose response is shorter than 32 tokens are discarded altogether. However, for multimodal data, many source prompts restrict the answer to a single word, a short phrase, or an option letter, and with the original prompts 12.5\% of the regenerated responses contain at most 8 tokens. We therefore replace such instructions with ``reason step by step'' or "explain each option", and append the same instruction to math problems and to questions from short-answer subsets such as chart, document, and table QA. Prompts that prescribe a specific output format, generation-style tasks such as captioning and OCR, and video samples are left unchanged, and the same rewriting is applied for every target model. For Qwen3.5-4B, 37.0\% of the prompts are rewritten. For the prompts whose answer-format instruction is replaced, the median response length grows from 3 to 130 tokens, and over the whole training set, the share of responses with at most 8 tokens drops from 12.5\% to 1.0\%.

\begin{table}[t]
  \centering
  \small
  \setlength{\tabcolsep}{1.5pt}
  \renewcommand{\arraystretch}{0.95}
  \begin{tabular*}{\linewidth}{@{\extracolsep{\fill}}ll*{5}{c}c@{}}
    \toprule
    \multirow{2}{*}{Task} & \multirow{2}{*}{Method} & \multicolumn{5}{c}{Concurrency} & \multirow{2}{*}{\textit{Avg.} $\tau$} \\
    \cmidrule(lr){3-7}
    & & 1 & 4 & 8 & 16 & 32 & \\
    \midrule
    \multicolumn{8}{@{}l}{\textbf{Qwen3.5-4B}} \\
    \midrule
    \multirow{3}{*}{CharXiv}
      & Baseline & 284 & 1022 & 1783 & 2690 & 3772 & -- \\
    \cmidrule(l){2-8}
      & \multirow{2}{*}{MMFlash} & 907 & 2555 & 3634 & 4590 & 5514 & \multirow{2}{*}{6.17} \\
      &                          & \spd{3.19} & \spd{2.50} & \spd{2.04} & \spd{1.71} & \spd{1.46} & \\
    \midrule
    \multirow{3}{*}{MMBench}
      & Baseline & 274 & 479 & 705 & 1051 & 1768 & -- \\
    \cmidrule(l){2-8}
      & \multirow{2}{*}{MMFlash} & 923 & 836 & 1347 & 1717 & 2859 & \multirow{2}{*}{5.74} \\
      &                          & \spd{3.37} & \spd{1.75} & \spd{1.91} & \spd{1.63} & \spd{1.62} & \\
    \midrule
    \multirow{3}{*}{DynaMath}
      & Baseline & 295 & 1052 & 1829 & 3199 & 4159 & -- \\
    \cmidrule(l){2-8}
      & \multirow{2}{*}{MMFlash} & 821 & 2430 & 3685 & 4705 & 5323 & \multirow{2}{*}{6.27} \\
      &                          & \spd{2.78} & \spd{2.31} & \spd{2.02} & \spd{1.47} & \spd{1.28} & \\
    \midrule
    \multicolumn{8}{@{}l}{\textbf{Qwen3.5-9B}} \\
    \midrule
    \multirow{3}{*}{ChartQA}
      & Baseline & 188 & 648 & 1241 & 1911 & 2133 & -- \\
    \cmidrule(l){2-8}
      & \multirow{2}{*}{MMFlash} & 701 & 1975 & 2795 & 3318 & 3500 & \multirow{2}{*}{6.17} \\
      &                          & \spd{3.72} & \spd{3.05} & \spd{2.25} & \spd{1.74} & \spd{1.64} & \\
    \midrule
    \multirow{3}{*}{MMStar}
      & Baseline & 190 & 700 & 1253 & 2080 & 3076 & -- \\
    \cmidrule(l){2-8}
      & \multirow{2}{*}{MMFlash} & 529 & 1663 & 2461 & 2893 & 3606 & \multirow{2}{*}{4.59} \\
      &                          & \spd{2.79} & \spd{2.38} & \spd{1.96} & \spd{1.39} & \spd{1.17} & \\
    \midrule
    \multirow{3}{*}{DynaMath}
      & Baseline & 189 & 708 & 1272 & 2128 & 3127 & -- \\
    \cmidrule(l){2-8}
      & \multirow{2}{*}{MMFlash} & 602 & 1860 & 2810 & 3604 & 4001 & \multirow{2}{*}{6.57} \\
      &                          & \spd{3.18} & \spd{2.63} & \spd{2.21} & \spd{1.69} & \spd{1.28} & \\
    \bottomrule
  \end{tabular*}
  \vspace{-3mm}
  \caption{Decode throughput (tok/s), speedup over the baseline, and average acceptance length ($\tau$) on vLLM under increasing serving concurrency, complementing Table~\ref{tab:concurrency_sglang}. For each MMFlash block the upper line reports throughput and the lower line the speedup at the same concurrency.}
  \label{tab:concurrency_vllm}
  \vspace{-4mm}
\end{table}

\subsection{Offline and Online Training}
\label{supp:training_details}

Training the drafter requires the hidden states of the target model on the training data, which can be obtained in two ways. Offline training runs the target model over the training set in advance and stores the tapped hidden states, which the drafter then reads during training. Online training instead runs the target model alongside the drafter, computing these hidden states for each batch on the fly and passing them directly to the drafter. Offline training computes the hidden states only once, but its storage grows with both the number of training tokens and the number of tapped layers, and typically reaches tens of terabytes for our data. We therefore adopt online training, which requires no extra storage at the cost of running the target model and transferring its tapped hidden states at every training step.

\subsection{Inference and Evaluation}
\label{supp:inference_eval}

For each benchmark, we randomly select 200 samples and measure the acceptance length and decoding speedup of all methods on this same subset.

\section{Additional Experimental Results}
\label{supp:additional_results}

\begin{table}[t]
  \centering
  \small
  \setlength{\tabcolsep}{1.5pt}
  \renewcommand{\arraystretch}{0.95}
  \begin{tabular*}{\linewidth}{@{\extracolsep{\fill}}l*{6}{c}@{}}
    \toprule
    \multirow{2}{*}{Setting}
      & \multicolumn{2}{c}{CharXiv} & \multicolumn{2}{c}{MMBench} & \multicolumn{2}{c}{DynaMath} \\
    \cmidrule(lr){2-3} \cmidrule(lr){4-5} \cmidrule(lr){6-7}
      & {\scriptsize $\tau$} & {\scriptsize Speedup} & {\scriptsize $\tau$} & {\scriptsize Speedup}
      & {\scriptsize $\tau$} & {\scriptsize Speedup} \\
    \midrule
    3-L & 4.91 & \spd{2.59} & 4.50 & \spd{2.48} & 5.34 & \spd{2.43} \\
    5-L & 5.39 & \spd{2.86} & 5.25 & \spd{2.58} & 5.75 & \spd{2.56} \\
    8-L & 5.59 & \spd{2.83} & 5.88 & \spd{2.67} & 5.94 & \spd{2.54} \\
    \bottomrule
  \end{tabular*}
  \vspace{-3mm}
  \caption{Effect of the number of drafter layers on MMFlash for Qwen3.5-4B at temperature $= 1$ (top-$p=0.95$, top-$k=20$), complementing Table~\ref{tab:drafter_scale}. $n$-L denotes a drafter with $n$ layers; all drafters are trained with block size 16 and read hidden features from 5 layers of the target model.}
  \label{tab:drafter_scale_t1}
  \vspace{-3mm}
\end{table}

This section reports the experimental results that do not fit in the main paper.

\subsection{Serving Concurrency on vLLM}
\label{supp:vllm}

Table~\ref{tab:concurrency_vllm} repeats the concurrency study of Table~\ref{tab:concurrency_sglang} on vLLM with the FlashAttention backend, keeping the benchmarks, prompts, generation cap, and MMFlash drafters unchanged; vLLM serves the drafter through its built-in DFlash speculator. MMFlash accelerates decoding on both target models at every concurrency level, by 1.17-3.72$\times$, with acceptance lengths close to those on SGLang. On Qwen3.5-4B, the speedups are generally lower than on SGLang because vLLM improves the baseline more than MMFlash: averaged over the seven benchmarks of Table~\ref{tab:speedup_qwen3}, vLLM raises the throughput of the baseline by 21-38\% and that of MMFlash by 17-28\% relative to SGLang. On Qwen3.5-9B, vLLM instead improves MMFlash more than the baseline at low concurrency, so the speedups at concurrency 1 and 4 exceed those on SGLang.

\begin{table}[t]
  \centering
  \small
  \setlength{\tabcolsep}{1.5pt}
  \renewcommand{\arraystretch}{0.95}
  \begin{tabular*}{\linewidth}{@{\extracolsep{\fill}}l*{6}{c}@{}}
    \toprule
    \multirow{2}{*}{Setting}
      & \multicolumn{2}{c}{CharXiv} & \multicolumn{2}{c}{MMBench} & \multicolumn{2}{c}{DynaMath} \\
    \cmidrule(lr){2-3} \cmidrule(lr){4-5} \cmidrule(lr){6-7}
      & {\scriptsize $\tau$} & {\scriptsize Speedup} & {\scriptsize $\tau$} & {\scriptsize Speedup}
      & {\scriptsize $\tau$} & {\scriptsize Speedup} \\
    \midrule
    5-H & 5.39 & \spd{2.86} & 5.25 & \spd{2.58} & 5.75 & \spd{2.56} \\
    8-H & 5.40& \spd{2.86} & 5.40 & \spd{2.59} & 5.83 & \spd{2.57} \\
    \bottomrule
  \end{tabular*}
  \vspace{-3mm}
  \caption{Effect of the number of tapped target hidden states on MMFlash for Qwen3.5-4B at temperature $= 1$, complementing Table~\ref{tab:target_taps}. $n$-H denotes a drafter that reads hidden features from $n$ layers of the target model.}
  \label{tab:target_taps_t1}
  \vspace{-4mm}
\end{table}

\subsection{Drafter Ablations at Temperature 1}
\label{supp:ablation_t1}

Tables~\ref{tab:drafter_scale_t1} and~\ref{tab:target_taps_t1} repeat the ablations on the number of drafter layers and on the number of target hidden states (Tables~\ref{tab:drafter_scale} and~\ref{tab:target_taps}) under sampling with temperature $T=1$, top-$p=0.95$, and top-$k=20$.

\section{Limitations \& Future Work}
\label{supp:limitations}

MMFlash is designed and evaluated for VLMs that take images or videos as input and generate text. For videos, it relies on a simple sparse attention pattern and lightweight fine-tuning rather than a drafter built specifically for video, and drafting on long videos remains harder than on images. A promising direction is therefore to tailor speculative decoding to videos, for example by selecting the drafter's visual context according to the question, compressing the redundancy across frames, and measuring visual dependence at the frame level so that training focuses on tokens grounded in specific moments. Another direction is to extend MMFlash to omni-modal models~\citep{Qwen2.5-Omni,qwen35omniblog}, which also take audio as input and generate streaming speech. This would require generalizing visual dependence to tokens that depend on audio or on joint audio-visual cues, handling even longer interleaved audio-visual contexts, and accelerating speech generation in addition to text.

\end{document}